\documentclass[10pt,conference]{IEEEtran}
\IEEEoverridecommandlockouts
\usepackage{cite}
\usepackage{amsmath,amssymb,amsfonts}
\usepackage{algorithmic}
\usepackage{graphicx}
\usepackage{textcomp}
\usepackage{xcolor}

\usepackage{xspace} 
\usepackage{tabularray}
\usepackage{graphicx}
\usepackage{textcomp}
\usepackage{multirow}
\usepackage{colortbl}
\usepackage{bbding}
\usepackage{makecell}
\usepackage{pifont}
\usepackage{color}
\usepackage{xcolor}
\usepackage{longtable}
\usepackage{booktabs}
\usepackage{hhline}
\usepackage{amssymb}
\usepackage{vcell}
\usepackage{subcaption}
\usepackage{tcolorbox}
\usepackage{verbatim}
\usepackage{float}
\usepackage{tcolorbox}
\usepackage{arydshln}

\newcommand{\codes}{\textsc{CodeS}\xspace}
\newcommand{\reposketcher}{RepoSketcher\xspace}
\newcommand{\filesketcher}{FileSketcher\xspace}
\newcommand{\sketchfiller}{SketchFiller\xspace}
\newcommand{\sketcheval}{SketchEval\xspace}
\newcommand{\sketchbleu}{SketchBLEU\xspace}

\newcommand{\nltorepo}{NL2Repo\xspace}
\newcommand{\readme}{\ding{192}$\mathrm{README.md}$}
\newcommand{\rs}{\ding{193}$\mathrm{Repository}$ $\mathrm{Sketch}$}
\newcommand{\fs}{\ding{194}$\mathrm{File}$ $\mathrm{Sketch}$}
\newcommand{\fb}{\ding{195}$\mathrm{Function}$ $\mathrm{Body}$}

\def\BibTeX{{\rm B\kern-.05em{\sc i\kern-.025em b}\kern-.08em
    T\kern-.1667em\lower.7ex\hbox{E}\kern-.125emX}}
\begin{document}

\title{\codes: Natural Language to Code Repository via\\ Multi-Layer Sketch}

\author{
\IEEEauthorblockN{1\textsuperscript{st} Daoguang Zan}
\thanks{The first four authors contributed equally to this work. Corresponding authors are Bei Guan, Zhiguang Yang, and Lizhen Cui.}
\IEEEauthorblockA{\textit{Institute of Software, Chinese Academy of Sciences} \\
\textit{University of Chinese Academy of Sciences}\\
Beijing, China \\
}
\and
\IEEEauthorblockN{2\textsuperscript{nd} Ailun Yu}
\IEEEauthorblockA{\textit{Peking University} \\
Beijing, China \\
}
\and
\IEEEauthorblockN{3\textsuperscript{rd} Wei Liu}
\IEEEauthorblockA{\textit{Peking University} \\
Beijing, China \\
}
\and
\IEEEauthorblockN{4\textsuperscript{th} Dong Chen}
\IEEEauthorblockA{\textit{Huawei Technologies Co., Ltd} \\
Beijing, China \\
}
\and
\IEEEauthorblockN{5\textsuperscript{th} Bo Shen}
\IEEEauthorblockA{\textit{Huawei Technologies Co., Ltd} \\
Beijing, China \\
}
\and
\IEEEauthorblockN{6\textsuperscript{th} Wei Li}
\IEEEauthorblockA{\textit{Chinese Academy of Sciences} \\
Beijing, China \\
}
\and
\IEEEauthorblockN{7\textsuperscript{th} Yafen Yao}
\IEEEauthorblockA{\textit{Huawei Technologies Co., Ltd} \\
Beijing, China \\
}
\and
\IEEEauthorblockN{8\textsuperscript{th} Yongshun Gong}
\IEEEauthorblockA{\textit{Shandong University} \\
Jinan, China \\
}
\and
\IEEEauthorblockN{9\textsuperscript{th} Xiaolin Chen}
\IEEEauthorblockA{\textit{Chinese Academy of Sciences} \\
Beijing, China \\
}
\and
\IEEEauthorblockN{10\textsuperscript{th} Bei Guan}
\IEEEauthorblockA{\textit{Chinese Academy of Sciences} \\
Beijing, China \\
}
\and
\IEEEauthorblockN{11\textsuperscript{th} Zhiguang Yang}
\IEEEauthorblockA{\textit{Funcun-wuyou Technologies Co., Ltd} \\
Beijing, China \\
}
\and
\IEEEauthorblockN{12\textsuperscript{th} Yongji Wang}
\IEEEauthorblockA{\textit{Chinese Academy of Sciences} \\
Beijing, China \\
}
\and
\IEEEauthorblockN{13\textsuperscript{th} Qianxiang Wang}
\IEEEauthorblockA{\textit{Huawei Technologies Co., Ltd} \\
Beijing, China \\
}
\and
\IEEEauthorblockN{14\textsuperscript{th} Lizhen Cui}
\IEEEauthorblockA{\textit{Shandong University} \\
Jinan, China \\
}
}

\maketitle

\begin{abstract}

The impressive performance of large language models (LLMs) on code-related tasks has shown the potential of fully automated software development.
In light of this, we introduce a new software engineering task, namely \textit{Natural Language to code Repository} (\nltorepo).
This task aims to generate an entire code repository from its natural language requirements.
To address this task, we propose a simple yet effective framework \codes, which decomposes \nltorepo into multiple sub-tasks by a multi-layer sketch.
Specifically, \codes includes three modules: \reposketcher, \filesketcher, and \sketchfiller.
\reposketcher first generates a repository's directory structure for given requirements;
\filesketcher then generates a file sketch for each file in the generated structure;
\sketchfiller finally fills in the details for each function in the generated file sketch.
To rigorously assess \codes on the \nltorepo task, we carry out evaluations through both automated benchmarking and manual feedback analysis.
For benchmark-based evaluation, we craft a repository-oriented benchmark, \sketcheval, and design an evaluation metric, \sketchbleu.
For feedback-based evaluation, we develop a VSCode plugin for \codes and engage $30$ participants in conducting empirical studies.
Extensive experiments prove the effectiveness and practicality of \codes on the \nltorepo task.
\end{abstract}


\section{Introduction}
\label{sec:introduction}

Large language models (LLMs) have achieved remarkable advancements in generating code based on natural language (NL2Code)~\cite{nl2code,codex,alphacode,codegen,starcoder,starcoder2,codellama,deepseekcoder,gpt4,claude3}. 
For instance, GPT-4 and Claude 3 can solve $67.0\%$ and $84.9\%$ of function-level Python programming tasks in a zero-shot manner.
Their impressive performance demonstrates the possibility of LLM-driven autonomous software development~\cite{alphaCodim}.
Hopefully, in the era of LLM, programming novices and even non-programmers will be able to build complex engineering projects through natural language.
In this sense, natural language is one step closer to being the ultimate ``programming language".
To explore this possibility, we propose a new software engineering task, called Natural Language to Code Repository (\nltorepo), 
with the goal of automatically generating an entire engineering code repository based on given natural language requirements.

However, automatically generating a real-world engineering project faces challenges 
due to the huge gap between the natural language description of requirements and the corresponding code repository.
This gap is reflected in two aspects:
(1)
A real-world engineering project is typically accompanied by detailed natural language requirements to describe its numerous functional features.
Unlike previous NL2Code tasks~\cite{nl2code}, natural language in \nltorepo will be longer and more complex.
Although code LLMs have shown excellent performance in translating natural language to function, their instruction following capability and long-context length currently may not be sufficient for the \nltorepo task.
(2)
A target code repository typically exhibits high structural integrity, both in its directory tree and source code.
Although code LLMs perform well in NL2Code, they excel more at generating simple standalone code snippets, such as line- or function-level code generation.
This is primarily because auto-regressive language models are inherently ill-suited for modeling structured information~\cite{autoreg}.

To fill this gap, one straightforward idea is to decompose the complex \nltorepo task into sub-tasks via a sketch-based approach~\cite{sketchconcept1,sketchconcept2,cert}.
Earlier studies have proved the effectiveness of sketching in function-level program synthesis under formal specifications, like program sketching~\cite{programSketching} and DreamCoder~\cite{dreamcoder}.
Nonetheless, those approaches fail in the \nltorepo task, as they do not support natural language input and hardly scale beyond function-level due to the large search space of candidate programs~\cite{bigspace,searchspace1,searchspace2}.
Notably, combining the sketch with code LLMs could complement both of these shortcomings, as probability-based models possess an inherent advantage at understanding natural language and handling large search space~\cite{searchspace1,searchspace2}.
Therefore, a thought-provoking question arises: how to leverage the insights of the sketch to empower code LLMs in better solving \nltorepo tasks?

To address this question, we propose \codes, a multi-layer \textbf{S}ketch-based framework for \nltorepo.
\codes includes three modules: \reposketcher, \filesketcher, and \sketchfiller, which divide the \nltorepo into three phases.
Specifically, \reposketcher first generates the repository's directory structure;
for each filename in structure, \filesketcher then generates its file sketch which omits function body details;
\sketchfiller finally fills in each function body.
The implementation of \codes framework involves two manners: prompt engineering and supervised fine-tuning. 
For the former, we directly utilize off-the-shelf code LLMs such as CodeLlama, DeepSeekCoder, and GPT-3.5, assigning them roles corresponding to the three modules.
Regarding the latter, we fine-tune these base models to improve their capability to generate entire code repositories, thereby enhancing the overall performance of \codes.

To thoroughly evaluate the solutions for \nltorepo task, we conduct both benchmark-based automated and feedback-based manual evaluation.
For benchmark-based evaluation, we craft a new benchmark, \sketcheval, by collecting $19$ latest GitHub repositories and devising a metric to assess the quality of generated repositories.
For feedback-based evaluation, we develop a VSCode plugin for it and invite $30$ participants to use it to implement two real-world engineering projects.
Extensive experiments on \sketcheval and empirical studies have proved the effectiveness and practicality of \codes.

Overall, our contributions can be summarized as follows:
\begin{itemize}
    \item We propose a new software engineering task, \nltorepo, with the goal of automatically generating an entire code repository based on given natural language requirements. 
    \item We introduce \codes, a sketch-based framework to solve \nltorepo task. It divides \nltorepo into three phases and tackles them layer by layer.
    \item To comprehensively evaluate the \nltorepo task, we craft a benchmark \sketcheval. It provides the evaluation of $19$ real-world GitHub repositories and introduces a repository-oriented evaluation metric \sketchbleu.
    Furthermore, we develop a VSCode plugin and conduct empirical studies.
    \item Extensive experiments on \sketcheval and empirical studies have proved the effectiveness and practicality of \codes. We have made our work publicly available\footnote{https://github.com/NL2Code/CodeS}.
\end{itemize}

\section{Framework of \codes}
\label{sec:framework}

\textbf{\nltorepo} tends to generate \emph{a functional code repository} from its \emph{natural language descriptions}. 
Though the input and output of a \nltorepo task are similar to a conventional NL2Code task~\cite{nl2code}, we consider it as a new task instead of a direct extension.
Even though both \nltorepo and NL2Code adopt natural language as their inputs. 
The inputs of \nltorepo tend to be much longer with diverse sources of information. 
It typically includes a project title, functional description, features, dependencies, usage examples, FAQs, etc.
Also, the functional descriptions in \nltorepo tend to be much more high-level compared to conventional NL2Code tasks.
The expected size of the produced code is way larger than its descriptions.
\nltorepo targets repository generation instead of code segments. It proposes new challenges in designing file structures, modules, and interfaces besides conventional code generation.
In this paper, we consider a project's README.md file as its natural language input. 
With this specific setting, we propose \codes, a multi-layer sketch framework for \nltorepo.

\begin{figure}[t]
    \small
    \centering
    \includegraphics[width=1.0\linewidth]{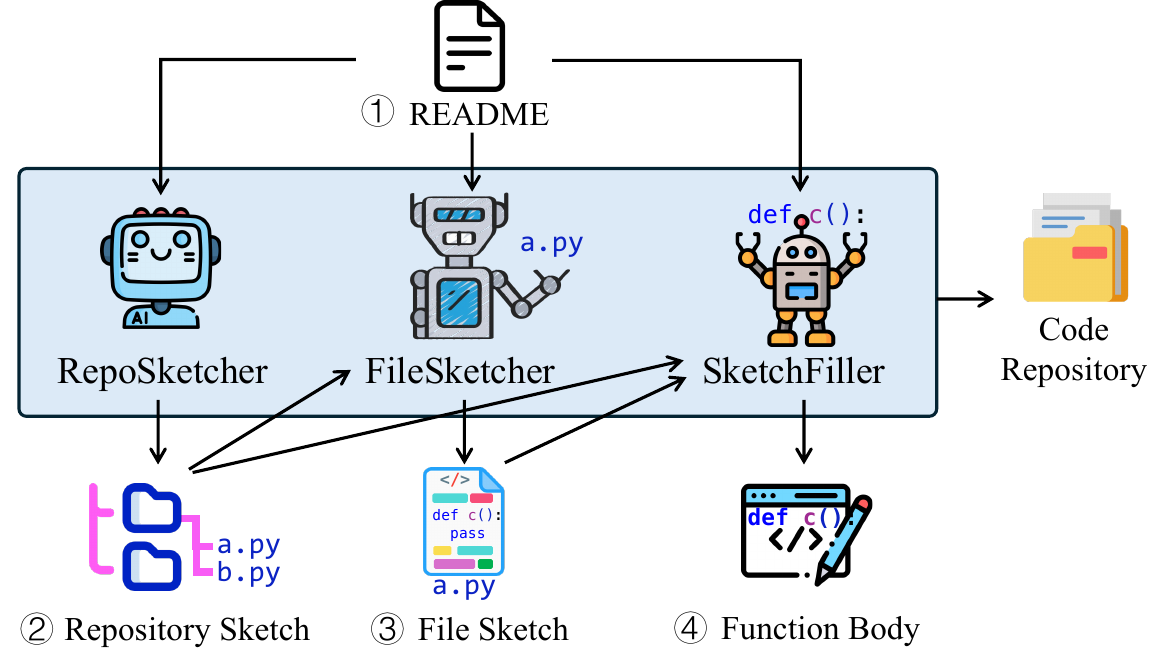}
    \caption{Overview of our proposed framework \codes.}
    \label{fig:codes_framework}
\end{figure}

\textbf{\codes} consists of three modules: \reposketcher, \filesketcher, and \sketchfiller, as Fig.~\ref{fig:codes_framework} shows.
Given a \readme, \reposketcher first aims to generate a \rs. It includes the names of all directories and files in the form of a tree.
Next, \filesketcher generates a \fs\ for each file in the generated repository sketch.
The file sketch outlines the import relations and the function definitions with empty bodies.
Finally, \sketchfiller is responsible for filling in the \fb\ for each function definition.
The arrows in Fig.~\ref{fig:codes_framework} show the information flow of \codes. It can be summarized as follows \ding{193}=$\mathrm{\reposketcher}$(\ding{192}), \ding{194}=$\mathrm{\filesketcher}$(\ding{192},\ding{193}), and \ding{195}=$\mathrm{\sketchfiller}$(\ding{192},\ding{193},\ding{194}).
Noted that, the \filesketcher and \sketchfiller aggregate information from previous stages.
This design enables \codes to progressively refine the sketch layer by layer without losing high-level task information.
The three modules of \codes are designed to collaborate seamlessly to address the requirements of the complex project.

\begin{figure*}[!t]
    \small
    \centering
    \includegraphics[width=0.975\linewidth]{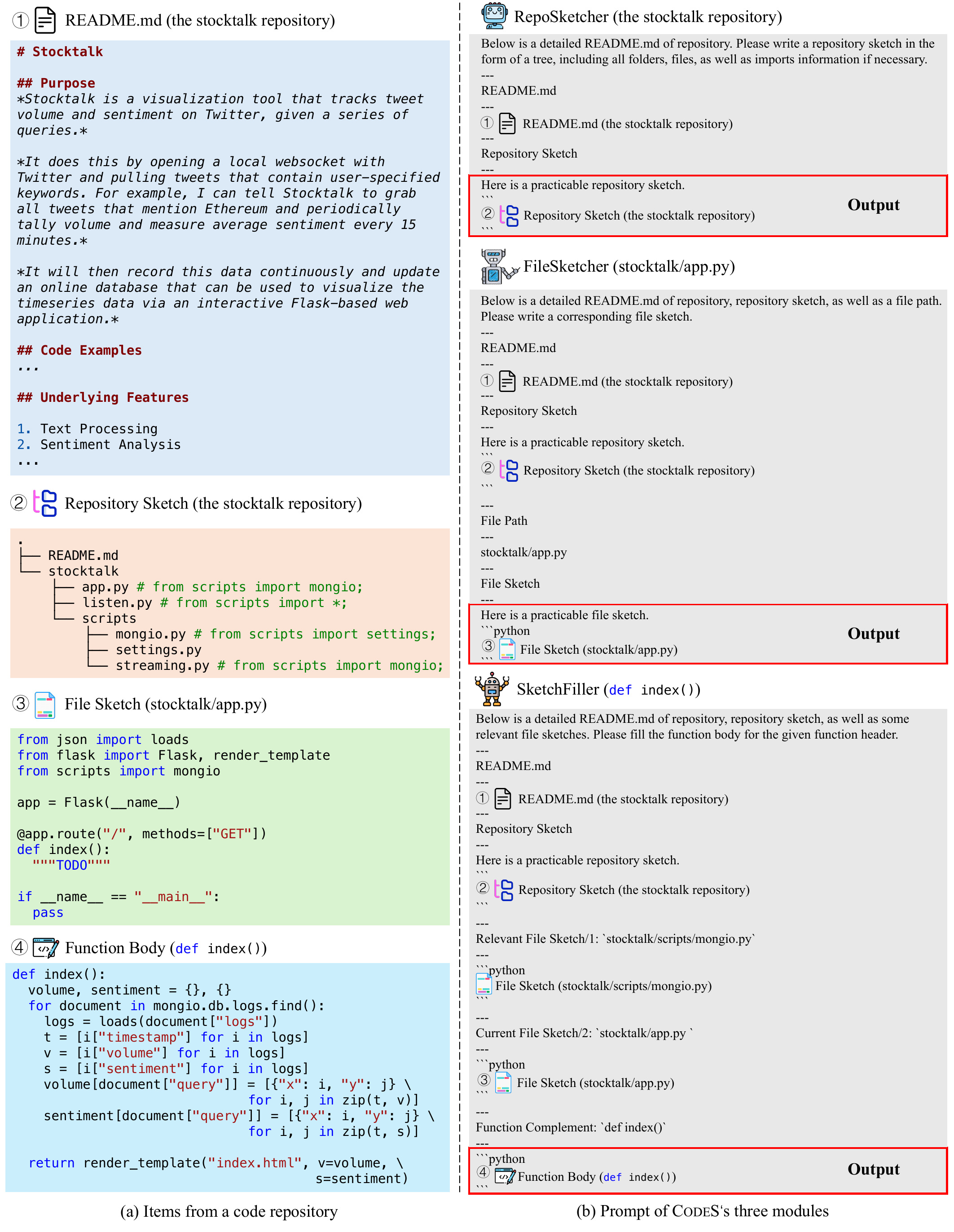}
    \caption{Prompt templates of \codes's three modules on a repository (https://github.com/anfederico/stocktalk).}
    \label{fig:codes_three_modules}
\end{figure*}

\section{Methodology}
\label{sec:methodology}

In practice, we implement the \codes framework for \nltorepo task in two manners, namely prompt engineering (PE) and supervised fine-tuning (SFT).
For the former, we prompt the off-the-shelf code LLMs, such as StarCoder2~\cite{starcoder2}, CodeLlama~\cite{codellama}, DeepSeekCoder~\cite{deepseekcoder}, and OpenAI's GPT-3.5~\cite{gpt35} to function as the three modules \reposketcher, \filesketcher and \sketchfiller.
For the latter, we leverage instruction fine-tuning to adapt base models to fulfill the functions of the three modules.
Both of the above implementation choices share the same workflow of \codes, with specified inputs and outputs of each module. The specification of inputs and outputs serves as the foundation for the construction of both the inference prompts and fine-tuning data.
Exemplary prompt templates of \codes are presented in Fig.~\ref{fig:codes_three_modules}. 
We proceed to elaborate on each module's design, explaining their input and output format.

Fig.~\ref{fig:codes_three_modules} (a) presents four items used in the inputs and outputs of three modules, including \readme, \rs, \fs\, and \fb.
A \readme\ contains the project title, description, features, installation, usage examples, table of contents, change logs, dependencies, FAQs, etc.
We only reserve the first five parts to prevent irrelevant or redundant information from affecting LLMs.
A \rs\ formats like a directory tree generated by the Linux \texttt{tree} command, but additionally attached with cross-file imports as annotations after each code file entry.
A \fs\ for each code file in the repository formats as a Python snippet where the bodies of all the functions are replaced with a placeholder statement.
A \fb\ is a function-level code snippet that completes the specified target function signature.
The above four items are used to compose the inputs and outputs of \codes' modules.

Fig.~\ref{fig:codes_three_modules} (b) lists the prompts used in each module, depicting their inputs and outputs.
The inputs of all the modules contain a brief description of the task and necessary items selected from \readme, \rs\, and \fs.
The outputs contain a prefixed statement identifying the response type, and a corresponding item of \rs, \fs\ or \fb\ respectively for the module of \reposketcher, \filesketcher, and \sketchfiller.
Specifically, \reposketcher accepts a \readme\ and outputs a \rs, while \filesketcher accepts a \readme\, a \rs\ and the target file path to generate, outputting a corresponding \fs.
For \sketchfiller, the input contains a \readme, a \rs, optional relevant \fs, current \fs\, and the target function signature to complete.
Among them, the relevant \fs\ corresponds to those dependent code files that the current file may import, hence they are included in the input for \sketchfiller's reference;
the current \fs\ presents both the context and location of the target function, with the body of the target function set as a ``TODO" placeholder statement while that of the other functions set as a ``pass" statement.
Finally, \sketchfiller outputs a corresponding \fb\ to complete the target function.
These above three modules are connected, with the output of one serving as the input of the next, collectively forming the \codes framework.

\section{\sketcheval: A Benchmark for \nltorepo}
\label{sec:sketcheval_a_benchmark_for_nl2repo}

We craft a benchmark \sketcheval to evaluate the \nltorepo tasks for Python.
\sketcheval includes: 
(1) a dataset of $19$ real-world repositories of varying complexity, which are delicately selected from the latest open-source GitHub projects.
(2) a metric to measure the similarity between two repositories, in terms of structures and semantics.

\subsection{Dataset for Evaluation}
\label{sec:repository_collection_of_benchmark}

To collect the evaluation repositories, we follow three steps: crawling, filtering, and grouping.
\textbf{Firstly}, we extensively crawl open-source repositories via GitHub API.
From GitHub, we request around $300$ most-starred repositories\footnote{https://api.github.com/search/repositories?q=language:Python+created:\textgreater20 23-08-01\&sort=stars\&order=desc\&per\_page=100\&page=\{page\}} with Python as the main language from August 2023 onwards\footnote{We choose this date to prevent data leakage in our base models.}.
\textbf{Secondly}, we refine the crawled repository set by filtering out repositories based on their type, quality, and topic.
(1) for types, we rule out those framework- or library-oriented repositories, mainly focusing on standalone projects.
(2) for quality, we select those repositories with a well-organized README and well-constructed source code.
For each repository, we require that its README document contains a meta description and a detailed feature description. The descriptions are necessary to declare the goal of a repository. we also check the validity of the source code by manually running the repository or reading the code.
(3) for topics, we try our best to cover diverse areas of repositories 
(e.g. AI, cyber security, shell tools, etc.).
\textbf{Thirdly}, we group the selected repositories into different difficulty levels based on the scale of a repository.
The repositories are grouped into three levels:
repositories of \textit{Hard} level should contain $\textgreater$ $10$ Python files or $\textgreater$ $2500$ Python code lines;
Repositories of \textit{Medium} level should contain $\textgreater$ $5$ Python files or $\textgreater$ $500$ Python code lines;
Other repositories are classified as \textit{Easy} level.

\begin{table}
\centering
\caption{Meta information of the selected repositories in \sketcheval.
\textbf{\#F.} and \textbf{\#L.} denotes the number of Python \underline{f}iles and code \underline{l}ines in the repository.
}
\resizebox{\linewidth}{!}{
\label{tab:repo_meta}
\begin{tabular}{cl|cccc} 
\toprule
\textbf{Difficulty}     & \textbf{Repository Name}      & \textbf{Stars} & \textbf{Created} & \textbf{\#F.} & \textbf{\#L.}  \\ 
\hline\hline
\multirow{5}{*}{Easy}   & CVE-2023-44487        & 205            & 2023-10-10       & 1              & 319             \\
                        & EVM\_inscription      & 227            & 2023-12-18       & 3              & 164             \\
                        & pitch-visualizer      & 192            & 2023-12-05       & 2              & 315             \\
                        & smol-podcaster        & 261            & 2023-08-08       & 1              & 277             \\
                        & web.Monitor           & 118            & 2023-09-09       & 1              & 237             \\ 
\hline
\multirow{8}{*}{Medium} & django-tui            & 212            & 2023-08-23       & 6              & 873             \\
                        & easier-docker         & 151            & 2023-11-23       & 10             & 274             \\
                        & epubhv                & 444            & 2023-09-04       & 7              & 958             \\
                        & every-breath-you-take & 541            & 2023-09-16       & 5              & 1497            \\
                        & fastui-chat           & 190            & 2023-12-17       & 8              & 367             \\
                        & kanban-python         & 228            & 2023-11-11       & 10             & 1945            \\
                        & libgen\_to\_txt       & 207            & 2023-10-16       & 9              & 570             \\
                        & van-gonography        & 384            & 2023-11-13       & 4              & 1030            \\ 
\hline
\multirow{6}{*}{Hard}   & EasyLiterature        & 174            & 2023-10-09       & 13             & 1564            \\
                        & flameshow             & 956            & 2023-09-24       & 19             & 2106            \\
                        & mactop                & 132            & 2023-12-05       & 43             & 3639            \\
                        & pygraft               & 611            & 2023-09-07       & 12             & 3882            \\
                        & pyobd                 & 709            & 2023-08-18       & 22             & 12220           \\
                        & sim-web-visualizer    & 223            & 2023-08-10       & 18             & 4870            \\
\bottomrule
\end{tabular}
}
\end{table}

As a result, we totally collect a set of $19$ repositories\footnote{https://github.com/NL2Code/CodeS/tree/main/validation/repos} as \textit{reference repositories}, including $5$ easy ones, $8$ medium ones, and $6$ hard ones.
The information of them is listed in Table~\ref{tab:repo_meta}.
For the three modules of \codes, these $19$ repositories provide $19$, $189$, and $1$,$166$ evaluation instances, respectively.

\subsection{Metric Definition}
\label{sec:metric_def}

To evaluate repository generation performance, we design a metric, namely \sketchbleu, which can calculate a repository-level similarity between the generated repository with the reference one.
We follow the idea of CodeBLEU~\cite{codebleu}, a metric that works effectively on function-level code evaluation.
CodeBLEU is a weighted combination of four parts including n-gram BLEU, weighted n-gram BLEU, syntactic tree match, and semantic data-flow match.
Based on CodeBLEU, we propose \sketchbleu by adapting the four parts of CodeBLEU to repository-level code evaluation, which can be defined as:
\begin{equation}\label{eq:sketch_bleu}
\begin{split}
    SketchBLEU &= \alpha\cdot BLEU'+\beta\cdot BLEU_{weight}'\\ &+\gamma\cdot Match_{struc}+\delta\cdot Match_{df}'
\end{split}
\end{equation}
where $BLEU'$ and $BLEU'_{weight}$ respectively denote the standard n-gram BLEU precision and the weighted n-gram BLEU precision that is calculated based on the concatenated source code of the repository;
$Match_{struc}$ denotes a n-hop tree match ($n=3$ in practice) of the \textit{structural tree}, which is constructed by concatenating the directory tree and the abstract syntax trees (ASTs) of Python code files in the repository;
$Match_{df}'$ denotes the semantic dataflow match between repositories, calculated by matching function-level dataflows between repositories via maximum weighted bipartite matching.
In our experiment, we assign equal weights to the four sub-parts of the metric, with $\alpha=\beta=\gamma=\delta=0.25$.
To transfer function-level CodeBLEU to repository-level SketchBLEU, we make efforts to adjust the four sub-parts of the original metric.
For instance, we adopt a less sensitive brevity penalty, because repositories implementing the same requirements can vary greatly in size.
Details of \sketchbleu can be found in our supplementary material (Section A).

\section{Research Questions}
\label{sec:research_questions}
We aim to comprehensively evaluate our framework \codes under the \nltorepo task by answering the following three Research Questions (RQs):

\begin{itemize}
    \item RQ1: How does \codes based on multi-layer sketch perform on the \nltorepo task?
    \item RQ2: What factors influence the performance of \codes?
    \item RQ3: Does \codes hold practical potential in real-world \nltorepo tasks?
\end{itemize}

We answer RQ1 by evaluating \codes against all its baselines on our constructed benchmark \sketcheval.
Regarding RQ2, we perform a detailed analysis for \codes from the perspectives of base model and instruction data.
In response to RQ3, we develop a VSCode plugin for \codes and invite $30$ participants to develop two projects using it.

\section{Experimental Setup}
\label{sec:experimental_setup}

\subsection{100 Code Repositories for Supervised Fine-Tuning}
\label{sec:code_repositories_for_sft}

To adapt base models for the \nltorepo task, we perform supervised fine-tuning based on a collection of $100$ code repositories.
To obtain these repositories, we first crawl Python repositories created before August 1, 2023 from GitHub.
Choosing this creation date as a filtering criterion aims to prevent data leakage into our newly constructed evaluation benchmark \sketcheval, thus ensuring the fairness of our experiments.
After that, we further exclude repositories with less than $100$ stars to ensure high-quality training data.

Based on the filtered repositories, we manually select $100$ high-quality Python ones\footnote{https://github.com/NL2Code/CodeS/tree/main/repos}. For each of these repositories, we extract its four parts: \ding{192}$\mathrm{README.md}$, \ding{193}$\mathrm{Repository}$ $\mathrm{Sketch}$, \ding{194}$\mathrm{File}$ $\mathrm{Sketch}$, and \ding{195}$\mathrm{Function Body}$.
One concrete example for these four parts is shown in Fig.~\ref{fig:codes_three_modules} (a).
Based on the collected repositories, we construct a total of $7$,$806$ instruction data for \codes's three modules: $100$ for \reposketcher, $1$,$191$ for \filesketcher, and $6$,$515$ for \sketchfiller.
Fig.~\ref{fig:codes_three_modules} showcases the prompt templates for \codes's three modules.

\subsection{\codes and Its Comparative Methods}

We implement the prompt engineering (PE) and supervised fine-tuning (SFT) versions of \codes based on multiple base models\footnote{The four base models were pre-trained with GitHub data before February 2023, September 2023, August 2023, and September 2021, respectively.}, including DeepSeekCoder-Instruct $7$\&$33$B~\cite{deepseekcoder}, StarCoder2 $3$\&$7$\&$15$B~\cite{starcoder2}, CodeLlama-Instruct $7$\&$13$\&$34$B~\cite{codellama}, and OpenAI's GPT-3.5-turbo~\cite{gpt35}.
Note that all GPT-3.5-based methods have only PE versions.
The baselines of \codes include two categories:
\begin{itemize}
    \item \textbf{Vanilla}: Our contributions lie in producing a complex code repository by utilizing a three-layer sketch. Therefore, a primary baseline for \codes is the method without any sketching, which means leveraging code LLMs to directly generate a repository.
    We implement the PE and SFT versions for it based on GPT-3.5 and CodeLlama.
    \item \textbf{Agent-based Methods}: 
    They were initially proposed to accomplish a specific objective through the collaborative working of multiple LLMs. 
    In the \nltorepo task with the objective of generating a repository, we reproduce three agent-based methods based on GPT-3.5 including ChatDev~\cite{chatdev}, AutoGPT~\cite{autogpt}, and AgentGPT~\cite{agentgpt}.
\end{itemize}

\subsection{Implementation Details}

We use DeepSpeed\footnote{https://github.com/microsoft/DeepSpeed} and Llama-Factory~\cite{llamafactory} to fine-tune base models.
During the instruction fine-tuning process of \codes, the batch size is set to $8$ per GPU card, gradient accumulation steps to $8$, and saving steps to $25$.
We set the learning rate to $5$e-$5$ with cosine decay, and enable fp16 to accelerate training.
AdamW~\cite{adamw} is used to optimize the parameters with $\beta$=$(0.9, 0.99)$ and $\epsilon$=$1$e-$8$.
Also, we employ Rotary Position Embedding (ROPE)~\cite{rope} to enhance the long-context modeling capability of the model.
For inference, we use greedy sampling across all our experiments on \sketcheval.
During the empirical study, we deploy \codes using vLLM~\cite{vllm} and provide API services for the VSCode plugin.
When requesting the API of \codes, we use nucleus sampling with the temperature of $0.2$, top\_p of $0.9$, frequency\_penalty of $0.35$, and presence\_penalty of $0.25$.
Regarding the instruction data, we use the \texttt{tree} command in Linux to extract repository sketch from a code repository.
We employ the \texttt{ast} library\footnote{https://docs.python.org/3/library/ast.html} for extracting the file sketch and function bodies from a Python file.
We format all the crawled source code using the \texttt{black} library\footnote{https://pypi.org/project/black}.
The version of OpenAI's GPT-3.5 used in our experiments is the \texttt{gpt-3.5-turbo-0613}\footnote{https://platform.openai.com/docs/models/gpt-3-5-turbo}.

\section{RQ1: The Effectiveness of \codes}
This section will initially evaluate the performance of \codes and its baselines on \sketcheval,
followed by delving into \codes's three modules.

\begin{table}[!t]
\caption{Performance of \codes and its baselines on \sketcheval. 
\textbf{E.}, \textbf{M.}, and \textbf{H.} stand for \underline{e}asy, \underline{m}edium, and \underline{h}ard.
$BLEU'$ (\textbf{B.}), $BLEU_{weight}'$ (\textbf{B.W.}), $Match_{struc}$ (\textbf{M.S.}), and $Match_{df}$ (\textbf{M.D.}) denote the four parts of \sketchbleu. DeepSeekCoder is abbreviated to DSCoder.}
\begin{center}
\scalebox{0.88}{
\rotatebox{0}{
\setlength{\tabcolsep}{2.0pt}
\begin{tabular}{l|l|lll:l|cccc} 
\toprule
\multirow{3}{*}{\textbf{Method}} & \multirow{3}{*}{\begin{tabular}[c]{@{}l@{}}\textbf{Base}\\\textbf{Model}\end{tabular}} & \multicolumn{8}{c}{\textbf{SketchEval}}                                                                                                                                                                                     \\ 
\cline{3-10}
                                 &                                                                                        & \textbf{E.(5)} & \textbf{M.(8)} & \textbf{H.(6)} & \textbf{All(19)} & \multicolumn{4}{c}{\textbf{All(19)}}                                                                                                                  \\ 
\cdashline{3-10}
                                 &                                                                                        & \multicolumn{4}{c|}{\textbf{SketchBLEU}}                            & \multicolumn{1}{l}{\textbf{B.}} & \multicolumn{1}{l}{\textbf{\textbf{B.W.}}} & \multicolumn{1}{l}{\textbf{M.S.}} & \multicolumn{1}{l}{\textbf{M.D.}}  \\ 
\hline\hline
\multicolumn{10}{c}{\textbf{\textcolor{red}{Prompt Engineering (PE)}}}                                                                                                                                                                                                                                                                                       \\ 
\hline
\multirow{4}{*}{Vanilla}         & GPT-3.5                                                                                & 19.23          & 14.56          & 12.82          & 15.24            & \multicolumn{1}{l}{14.25}       & \multicolumn{1}{l}{14.11}                  & \multicolumn{1}{l}{18.41}         & \multicolumn{1}{l}{14.19}          \\
                                 & CodeLlama 7B                                                                           & 18.45          & 15.63          & 9.54           & 14.45            & \multicolumn{1}{l}{13.55}       & \multicolumn{1}{l}{13.74}                  & \multicolumn{1}{l}{17.36}         & \multicolumn{1}{l}{13.15}          \\
                                 & CodeLlama 13B                                                                          & 20.63          & 15.48          & 12.33          & 15.84            & \multicolumn{1}{l}{14.88}       & \multicolumn{1}{l}{15.59}                  & \multicolumn{1}{l}{19.42}         & \multicolumn{1}{l}{13.47}          \\
                                 & CodeLlama 34B                                                                          & 23.52          & 17.36          & 11.12          & 17.01            & \multicolumn{1}{l}{15.63}       & \multicolumn{1}{l}{16.73}                  & \multicolumn{1}{l}{22.57}         & \multicolumn{1}{l}{13.11}          \\ 
\hdashline
ChatDev                          & GPT-3.5                                                                                & 56.13          & 46.93          & 31.17          & 45.13            & 41.06                           & 42.44                                      & 52.35                             & 44.65                              \\
AutoGPT                          & GPT-3.5                                                                                & 55.53          & 45.24          & 31.76          & 43.69            & 41.34                           & 43.86                                      & 51.52                             & 38.04                              \\
AgentGPT                         & GPT-3.5                                                                                & 53.35          & 44.14          & 30.13          & 42.14            & 40.01                           & 39.64                                      & 53.34                             & 35.57                              \\ 
\hdashline
\multirow{9}{*}{\codes}           & GPT-3.5                                                                                & 52.41          & 52.96          & 36.54          & 47.63            & 44.27                           & 44.58                                      & 57.21                             & 44.44                              \\
                                 & DSCoder 7B                                                                             & 50.15          & 52.24          & 34.29          & 46.02            & 42.56                           & 43.00                                      & 46.61                             & 52.91                              \\
                                 & DSCoder 33B                                                                            & \textbf{58.47} & \textbf{57.36} & \textbf{39.26} & \textbf{51.94}   & \textbf{46.14}                  & \textbf{48.64}                             & \textbf{59.33}                    & 53.65                              \\
                                 & StarCoder2 3B                                                                          & 28.87          & 28.24          & 14.53          & 24.08            & 20.56                           & 20.24                                      & 32.51                             & 23.01                              \\
                                 & StarCoder2 7B                                                                          & 32.45          & 29.95          & 15.24          & 25.96            & 23.24                           & 24.94                                      & 31.23                             & 24.43                              \\
                                 & StarCoder2 15B                                                                         & 35.13          & 32.54          & 18.35          & 28.74            & 25.35                           & 27.75                                      & 35.84                             & 26.02                              \\
                                 & CodeLlama 7B                                                                           & 50.63          & 51.52          & 33.47          & 45.59            & 42.45                           & 43.03                                      & 45.58                             & 51.30                              \\
                                 & CodeLlama 13B                                                                          & 51.44          & 52.15          & 34.64          & 46.43            & 41.14                           & 42.52                                      & 46.57                             & 55.49                              \\
                                 & CodeLlama 34B                                                                          & 56.14          & 54.50          & 36.74          & 49.32            & 43.55                           & 44.51                                      & 52.73                             & \textbf{56.49}                     \\ 
\hline
\multicolumn{10}{c}{\textbf{\textcolor{red}{Supervised Fine-Tuning (SFT)}}}                                                                                                                                                                                                                                                                                   \\ 
\hline
\multirow{3}{*}{Vanilla}         & CodeLlama 7B                                                                           & 26.35          & 22.53          & 21.59          & 23.24            & \multicolumn{1}{l}{22.96}       & \multicolumn{1}{l}{22.46}                  & \multicolumn{1}{l}{25.53}         & \multicolumn{1}{l}{22.01}          \\
                                 & CodeLlama 13B                                                                          & 30.14          & 26.56          & 22.21          & 26.13            & \multicolumn{1}{l}{25.31}       & \multicolumn{1}{l}{26.23}                  & \multicolumn{1}{l}{28.87}         & \multicolumn{1}{l}{24.11}          \\
                                 & CodeLlama 34B                                                                          & 35.63          & 29.44          & 27.48          & 30.45            & \multicolumn{1}{l}{28.46}       & \multicolumn{1}{l}{28.78}                  & \multicolumn{1}{l}{33.77}         & \multicolumn{1}{l}{30.79}          \\ 
\hdashline
\multirow{8}{*}{\codes}           & DSCoder 7B                                                                             & 57.47          & 57.78          & 52.86          & 56.14            & 54.35                           & 54.75                                      & 65.66                             & 49.80                              \\
                                 & DSCoder 33B                                                                            & 65.05          & 63.36          & \textbf{61.53} & 63.23            & \textbf{60.44}                  & \textbf{62.41}                             & 68.35                             & 61.72                              \\
                                 & StarCoder2 3B                                                                          & 35.56          & 32.03          & 24.57          & 30.60            & 26.36                           & 28.76                                      & 38.52                             & 28.76                              \\
                                 & StarCoder2 7B                                                                          & 48.23          & 46.31          & 39.78          & 44.75            & 41.45                           & 42.32                                      & 42.65                             & 52.58                              \\
                                 & StarCoder2 15B                                                                         & 57.25          & 55.98          & 47.56          & 53.66            & 49.99                           & 51.46                                      & 58.35                             & 54.84                              \\
                                 & CodeLlama 7B                                                                           & 55.96          & 56.11          & 50.78          & 54.39            & 50.49                           & 51.81                                      & 63.92                             & 49.35                              \\
                                 & CodeLlama 13B                                                                          & 58.24          & 57.25          & 56.35          & 57.23            & 54.24                           & 53.26                                      & 58.25                             & 63.17                              \\
                                 & CodeLlama 34B                                                                          & \textbf{65.14} & \textbf{64.24} & 60.56          & \textbf{63.31}   & 57.46                           & 59.25                                      & \textbf{71.55}                    & \textbf{64.98}                     \\
\bottomrule
\end{tabular}
}
}
\label{tab:main_results}
\end{center}
\end{table}

\subsection{Performance Analysis of \codes}

Table~\ref{tab:main_results} shows the performance of \codes and its baselines on \sketcheval.
Compared to the baseline Vanilla without sketching, \codes with multi-layer sketch exhibits significant performance improvements.
This underscores the necessity of incorporating the multi-layer sketch into code LLMs.
Additionally, \codes outperforms three agent-based baselines in medium and hard repositories, while performing comparably in easy ones.
This is because easy repositories do not heavily rely on the multi-layer sketch.
Conversely, in hard repositories with complex directory structures, the baselines will fall short because they do not explicitly decompose repositories' complexity, highlighting the effectiveness of \codes.
Furthermore, we observe that \codes's SFT versions consistently surpass the PE ones.
As an example, after supervised fine-tuning, \codes based on CodeLlama $7$B, $13$B, and $34$B increases \sketchbleu scores from $45.59\%$, $46.43\%$, $49.32\%$ to $54.39\%$, $57.23\%$, $63.31\%$, respectively.
This observation emphasizes the necessity of supervised fine-tuning on our crafted instruction data in Section~\ref{sec:code_repositories_for_sft}.
Besides, for all PE versions of \codes, there exists a notable performance drop as the difficulty of the repositories increases.
However, the SFT versions show a less pronounced decline, further underscoring the value of instruction fine-tuning.

\begin{figure}[t]
    \small
    \centering
    \includegraphics[width=1.0\linewidth]{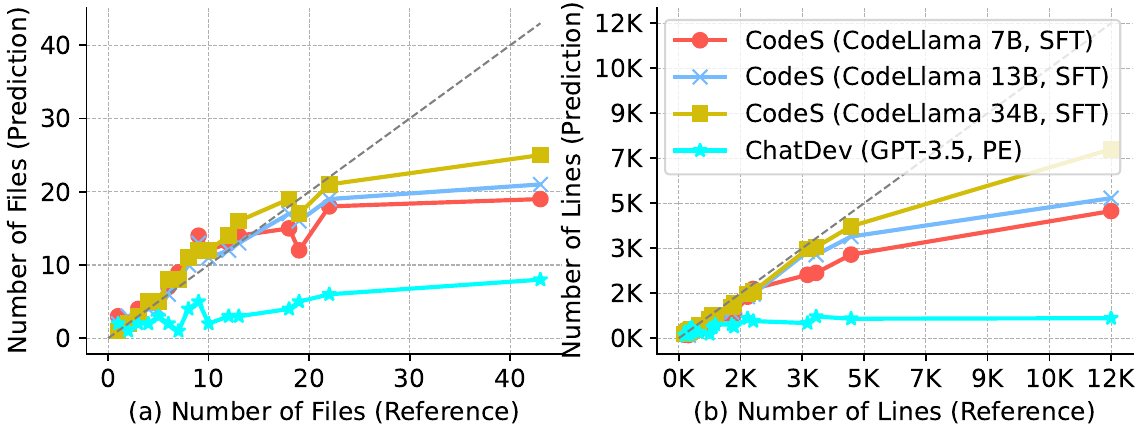}
    \caption{Comparison between the predicted repositories and the reference ones in terms of number of files (a) and lines (b).}
    \label{fig:gt_files_lines_vs_generated_files_lines}
\end{figure}

In addition, we also compare the number of files/lines in predicted code repositories versus that in reference ones.
The results of \codes and ChatDev are displayed in Fig.~\ref{fig:gt_files_lines_vs_generated_files_lines}.
Regardless of the number of files or lines, \codes's predictions are consistently closer to the reference, compared to that of ChatDev.
This suggests that \codes is able to predict the number of files and lines of the target repository more accurately.
Interestingly, regardless of the complexity of requirements, the repositories generated by ChatDev always contain no more than $8$ files or $842$ lines.
\codes, in contrast, supports up to $25$ files and $7$,$334$ lines.
This indicates that \codes, with its multi-layer sketch, is able to mitigate the shortcomings of code LLMs in modeling long contexts.

\subsection{Performance Analysis of Three Modules of \codes}

Table~\ref{tab:three_phase_results} shows the performance of \codes's three modules on code repositories of varying difficulty.
In this experiment, all input items of each module are derived from the reference repository.
For the PE version of \codes, as the difficulty increases, the most significant performance drop occurs during the first of the three phases.
For example, CodeLlama 13B experiences a $28\%$ decrease in BLEU score from easy ($77\%$) to hard levels ($49\%$).
This suggests that the primary obstacle for the PE version of \codes lies in the first (generating repository sketch) phase when addressing the complex repository generation.
Upon applying SFT, \codes demonstrates a significant performance improvement in the first phase.
Such findings further highlight the importance of instruction fine-tuning.
Table~\ref{tab:three_phase_statistic} offers a statistical analysis of the outputs from \codes's three modules, drawing conclusions similar to those in Table~\ref{tab:three_phase_results}.
For instance, for the PE version, the average number of files in the generated repository sketch for CodeLlama $34$B is $5$, while the reference is $10$. 
However, after SFT, it reaches an average of $11$ files, further demonstrating the potential of our curated instruction data.

\begin{table}[!t]
\caption{Performance of \reposketcher (phase 1), \filesketcher (phase 2), and \sketchfiller (phase 3) of \codes on \sketcheval's code repositories of varying difficulty. 
\textbf{E.}, \textbf{M.}, and \textbf{H.} denotes \underline{e}asy, \underline{m}edium, and \underline{h}ard.
}
\begin{center}
\scalebox{0.98}{
\rotatebox{0}{
\setlength{\tabcolsep}{2.0pt}
\begin{tabular}{l|l|lllc|lllc|lllc} 
\toprule
\multirow{3}{*}{\textbf{Method}} & \multirow{3}{*}{\begin{tabular}[c]{@{}l@{}}\textbf{\textbf{Base}}\\\textbf{\textbf{Model}}\end{tabular}} & \multicolumn{4}{c|}{\textbf{Phase 1}}                                                             & \multicolumn{4}{c|}{\textbf{\textbf{Phase 2}}}                                                   & \multicolumn{4}{c}{\textbf{\textbf{Phase 3}}}                                                       \\
                                 &                                                                                                          & \multicolumn{4}{c|}{\begin{tabular}[c]{@{}c@{}}\textbf{Repo. Sketch}\\\textbf{BLEU}\end{tabular}} & \multicolumn{4}{c|}{\begin{tabular}[c]{@{}c@{}}\textbf{File Sketch}\\\textbf{BLEU}\end{tabular}} & \multicolumn{4}{c}{\begin{tabular}[c]{@{}c@{}}\textbf{Fun. Body}\\\textbf{CodeBLEU}\end{tabular}}  \\ 
\cdashline{3-14}
                                 &                                                                                                          & \textbf{E.} & \textbf{M.} & \textbf{H.} & \textbf{All}                                            & \textbf{E.} & \textbf{M.} & \textbf{H.} & \textbf{All}                                           & \textbf{E.} & \textbf{M.} & \textbf{H.} & \textbf{All}                                              \\ 
\hline
\multicolumn{14}{c}{\textbf{\textcolor{red}{Prompt Engineering (PE)}}}                                                                                                                                                                                                                                                                                                                                                                                        \\ 
\hline
\multirow{4}{*}{\codes}           & GPT-3.5                                                                                                  & 75          & 69          & 45          & 63                                                      & 47          & 45          & 39          & 45                                                     & 66          & 64          & 62          & 64                                                        \\
                                 & CodeLlama 7B                                                                                             & 67          & 64          & 46          & 59                                                      & 45          & 38          & 38          & 40                                                     & 65          & 63          & 64          & 64                                                        \\
                                 & CodeLlama 13B                                                                                            & 77          & 72          & 49          & 66                                                      & 45          & 43          & 44          & 44                                                     & 71          & 67          & 67          & 68                                                        \\
                                 & CodeLlama 34B                                                                                            & \textbf{84} & \textbf{81} & \textbf{60} & \textbf{75}                                             & \textbf{56} & \textbf{53} & \textbf{50} & \textbf{53}                                            & \textbf{76} & \textbf{70} & \textbf{71} & \textbf{72}                                               \\ 
\hline
\multicolumn{14}{c}{\textbf{\textcolor{red}{Supervised Fine-Tuning (SFT)}}}                                                                                                                                                                                                                                                                                                                                                                                   \\ 
\hline
\multirow{3}{*}{\codes}           & CodeLlama 7B                                                                                             & 72          & 70          & 62          & 68                                                      & 55          & 50          & 52          & 52                                                     & 74          & 71          & 68          & 71                                                        \\
                                 & CodeLlama 13B                                                                                            & 81          & 81          & 65          & 76                                                      & 62          & 57          & 56          & 58                                                     & 80          & 76          & 79          & 78                                                        \\
                                 & CodeLlama 34B                                                                                            & \textbf{88} & \textbf{84} & \textbf{68} & \textbf{80}                                             & \textbf{64} & \textbf{61} & \textbf{58} & \textbf{61}                                            & \textbf{83} & \textbf{81} & \textbf{79} & \textbf{81}                                               \\
\bottomrule
\end{tabular}
}
}
\label{tab:three_phase_results}
\end{center}
\end{table}

\begin{table}[!t]
\caption{Summary statistics for the outputs of \reposketcher, \filesketcher, and \sketchfiller. \textbf{\#D.}, \textbf{\#F.}, and \textbf{\#l.} are the number of \underline{d}irectories, \underline{f}iles, and \underline{i}mports in repository sketch; \textbf{\#c.} and \textbf{\#f.} represent \underline{c}lass and \underline{f}unction counts in file sketch; \textbf{\#API} and \textbf{\#L.} denote \underline{API} and \underline{l}ine counts in function body.}
\begin{center}
\scalebox{0.98}{
\rotatebox{0}{
\setlength{\tabcolsep}{3.0pt}
\begin{tabular}{l|l|ccc|cc|cc} 
\toprule
\multirow{2}{*}{\textbf{Method}} & \multirow{2}{*}{\begin{tabular}[c]{@{}l@{}}\textbf{Base}\\\textbf{Model}\end{tabular}} & \multicolumn{3}{c|}{\begin{tabular}[c]{@{}c@{}}\textbf{Phase 1}\\\textbf{Repo. Sketch}\end{tabular}} & \multicolumn{2}{c|}{\begin{tabular}[c]{@{}c@{}}\textbf{\textbf{Phase 2}}\\\textbf{\textbf{File Sketch}}\end{tabular}} & \multicolumn{2}{c}{\begin{tabular}[c]{@{}c@{}}\textbf{\textbf{\textbf{\textbf{Phase 3}}}}\\\textbf{\textbf{\textbf{\textbf{Fun. Body}}}}\end{tabular}}  \\
                                 &                                                                                        & \textbf{\#D.} & \textbf{\textbf{\#F.}} & \textbf{\#l.}                                               & \textbf{\textbf{\#c}} & \#\textbf{\textbf{\textbf{\textbf{f.}}}}                                                      & \textbf{\#API} & \textbf{\#L.}                                                                                                                          \\ 
\hline\hline
Reference                        & -                                                                                      & 4             & 10                     & 7                                                           & 2                     & 5                                                                                             & 6              & 13                                                                                                                                     \\ 
\hline
\multicolumn{9}{c}{\textcolor{red}{\textbf{Prompt Engineering (PE)}}}                                                                                                                                                                                                                                                                                                                                                                                                                                                  \\ 
\hline
\multirow{4}{*}{\codes}           & GPT-3.5                                                                                & 2             & 3                      & 2                                                           & 1                     & 4                                                                                             & 5              & 10                                                                                                                                     \\
                                 & CodeLlama 7B                                                                           & 1             & 4                      & 2                                                           & 0                     & 3                                                                                             & 3              & 12                                                                                                                                     \\
                                 & CodeLlama 13B                                                                          & 2             & 4                      & 4                                                           & 1                     & 3                                                                                             & 4              & 13                                                                                                                                     \\
                                 & CodeLlama 34B                                                                          & 3             & 5                      & 5                                                           & 1                     & 4                                                                                             & 6              & 12                                                                                                                                     \\ 
\hline
\multicolumn{9}{c}{\textcolor{red}{\textbf{Supervised Fine-Tuning (SFT)}}}                                                                                                                                                                                                                                                                                                                                                                                                                                               \\ 
\hline
\multirow{3}{*}{\codes}           & CodeLlama 7B                                                                           & 2             & 10                     & 4                                                           & 1                     & 3                                                                                             & 4              & 13                                                                                                                                     \\
                                 & CodeLlama 13B                                                                          & 3             & 10                     & 5                                                           & 2                     & 5                                                                                             & 4              & 12                                                                                                                                     \\
                                 & CodeLlama 34B                                                                          & 4             & 11                     & 8                                                           & 1                     & 6                                                                                             & 6              & 14                                                                                                                                     \\
\bottomrule
\end{tabular}
}
}
\label{tab:three_phase_statistic}
\end{center}
\end{table}

\section{RQ2: Influencing Factors of \codes}
In this section, we explore the factors that influence the performance of \codes, focusing on two critical aspects: the base model and instruction data. 
Specifically, we first investigate the impact of different sizes of various base models on both the PE and SFT versions. 
Then, we study how dataset scale and instruction format affect the effectiveness of SFT.

\subsection{Base Model}

\begin{figure}[t]
    \small
    \centering
    \includegraphics[width=1.0\linewidth]{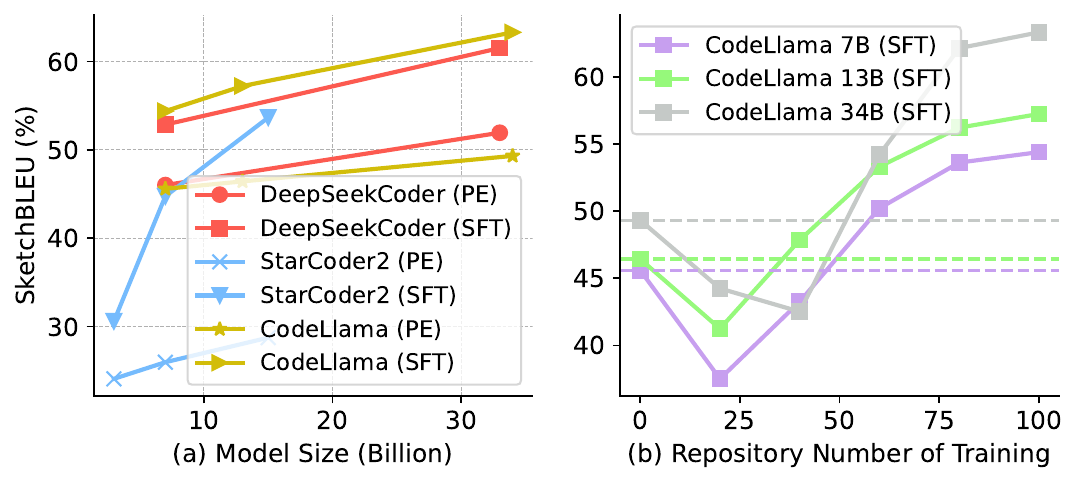}
    \caption{\sketchbleu of \codes across different settings on \sketcheval.}
    \label{fig:various_models}
\end{figure}

Fig.~\ref{fig:various_models} (a)  illustrates the \sketchbleu of \codes's PE and SFT versions across various base models of different sizes.
Generally, a larger base model yields a higher \sketchbleu for both the PE and SFT versions, and SFT consistently outperforms PE by an average improvement of $13.15\%$ in \sketchbleu .
Furthermore, the effectiveness of instruction fine-tuning appears to become increasingly significant as the model size grows.
For example, for CodeLlama with a size of $7$B, the improvement on \sketchbleu brought by SFT compared with PE is $8.89\%$, while for CodeLlama of larger size, such as $33$B, the improvement in \sketchbleu jumps to $13.99\%$.
This is mainly because the effectiveness of SFT depends largely on the knowledge encoded in LLMs~\cite{scaling}. 
As the model size increases, its ability to encode and leverage intricate knowledge also expands, consequently boosting the effectiveness of instruction fine-tuning that unlocks the potential of LLMs for \nltorepo tasks.

\subsection{Instruction Data}

\subsubsection{Data scale}
Fig.~\ref{fig:various_models} (b) shows the performance of \codes on \sketcheval as the scale of the instruction dataset expands.
It demonstrates that SFT becomes effective when the number of repositories used for fine-tuning surpasses a specific threshold. 
As can be seen from Fig.~\ref{fig:various_models} (b), this threshold for CodeLlama series models is $60$.
In other words, the instruction fine-tuning on CodeLlama outperforms prompt engineering (i.e., \#Repositories = $0$) only when the number of repositories for fine-tuning equals or exceeds $60$.
When the number of repositories falls below the specific threshold, SFT results in a decline in performance, which is particularly noticeable with CodeLlama $7$B. 
This decline is mainly attributed to the over-fitting caused by a relatively small fine-tuning dataset scale.
When the number of repositories exceeds the specific threshold, \sketchbleu tends to increase as more repositories are involved in fine-tuning. However, this increase is limited; there is a tendency to reach a plateau when the number of training repositories reaches or exceeds $80$.
These observations further validate our decision to use $100$ repositories for fine-tuning, which achieves a cost-effective balance between model performance and fine-tuning cost.

\begin{table}[!t]
\caption{Ablation study of \codes.}
\begin{center}
\scalebox{1.0}{
\rotatebox{0}{
\setlength{\tabcolsep}{3.5pt}
\begin{tabular}{l|llll} 
\toprule
\multirow{2}{*}{\textbf{Method}} & \multicolumn{4}{c}{\textbf{SketchEval}}                                          \\
                                 & \textbf{Easy (5)} & \textbf{Middle (8)} & \textbf{Hard (6)} & \textbf{All (19)}  \\ 
\hline\hline
\codes (CodeLlama 13B)            & \textbf{58.24}             & \textbf{57.25}               & \textbf{56.35}             & \textbf{57.23}              \\ 
\hline
~ +Table of Contents             & 57.56             & 55.00               & 50.53             & 54.26              \\
~ +Change Logs                   & 56.14             & 53.24               & 51.03             & 53.31              \\
~ +Dependencies                  & 58.02             & 54.89               & 49.42             & 53.99              \\
~ +FAQs                          & 57.47             & 56.35               & 50.46             & 54.78              \\ 
\cdashline{1-1}
~ -Import                        & 56.35             & 50.01               & 44.04             & 49.79              \\
~ -Tree Format                   & 56.63             & 53.45               & 48.87             & 52.84              \\
~ -TODO                          & 53.15             & 49.13               & 42.41             & 48.07              \\
~ -Relevant File Sketch          & 57.94             & 52.56               & 43.57             & 51.14              \\
\bottomrule
\end{tabular}
}
}
\label{tab:ablation_study}
\end{center}
\end{table}

\subsubsection{Instruction Format}
Table~\ref{tab:ablation_study} exhibits the results of the ablation study conducted on the components of \codes.
Specifically, we propose eight variants of \codes's prompt by adding (+) new components or removing (-) existing components.
For \readme, \codes employs a data pre-processing process to filter out irrelevant parts, including the table of contents, change logs, dependencies, and FAQs. In Table~\ref{tab:ablation_study}, we separately add each of these four filtered parts to analyze their influence on the performance of \codes.
Regarding \rs, it includes relevant ``import" information appended to each filename and adopts a tree format to organize the repository sketch. To evaluate the effect of these two components, we (1) remove the import information (-Import) and (2) replace the tree format with a multi-layer parenthesis to describe the file structure (-Tree Format).
As for \fs, it utilizes a ``TODO" keyword as a placeholder to enable the model to identify completion places, 
and we also remove it (-TODO) to analyze \codes.
Additionally, in the prompt for the \sketchfiller module, \codes includes a part for relevant file sketch to capture cross-file information. In Table~\ref{tab:ablation_study}, we remove this part (-Relevant File Sketch) to investigate the impact of cross-file information on \codes.

Table~\ref{tab:ablation_study} shows that \codes consistently achieves the highest \sketchbleu, indicating the superiority of its instruction fine-tuning strategy.
Additionally, it can be observed that this superiority becomes more significant as task complexity increases.
For example, after removing the import information, the decrease in \sketchbleu on the easy task is $1.89\%$, while that value on the hard task is $12.31\%$.
The same phenomenon can be observed when Relevant File Sketch is removed.
This discrepancy is mainly attributed to the augmented complexity of file dependencies in harder tasks. 
For all added components, we observe that meaningless noise information in \ding{192}$\mathrm{README.md}$ degrades the model's performance. Therefore, we can conclude that the pre-processing of \ding{192}$\mathrm{README.md}$ is a necessary step for enhancing the performance for \nltorepo tasks.
Among the existing components in \codes, the removal of the ``TODO" keyword results in the largest decrease in \sketchbleu, emphasizing the pivotal role of this keyword in enabling LLMs to identify the appropriate generation context.
Regarding the removal of the tree format, the pre-trained model is more familiar with this format instead of others such as multi-layer parenthesis since the tree format is a widely adopted pattern in practice.
Therefore, the fine-tuning based on the tree format is more effective since it is aligned with the knowledge in LLMs.

\section{RQ3: The Practicality of \codes}
\label{sec:rq3_the_practicality_of_codes}
In light of its superior performance on our crafted benchmark \sketcheval, we are keen to delve into the practicality of \codes in boosting programming efficiency.
We thus develop ``\codes Extended'', a VSCode extension based on \codes.
To evaluate its practicality, we invite $30$ participants to use \codes Extended and other competitive code assistants to implement two real-world engineering projects.
We monitor their entire implementation process to analyze the advantages and disadvantages of different code assistants.
This section first introduces all the code assistants participating in the evaluation, then describes the two projects that need to be implemented, followed by the procedure of our empirical study, and finally presents the evaluation results.

\subsection{Code Assistants}
\label{sec:code_assistants}
\begin{itemize}
    \item \textbf{Python Extended}\footnote{https://github.com/tushortz/vscode-Python-Extended}: 
    This VSCode extension employs a rule-based strategy to enhance Python programming by providing API-level code suggestions.
    \item \textbf{GitHub Copilot}\footnote{https://github.com/features/copilot}: It is a VSCode extension driven by LLMs, co-developed by GitHub and OpenAI. It can improve coding efficiency by automatically generating line-level or function-level code snippets based on the context of the repository.
    \item \textbf{ChatDev Extended}: We deploy ChatDev~\cite{chatdev} to implement this VSCode extension. It is capable of generating an entire code repository by allowing GPT-3.5 to play different roles (e.g., CEO, CTO, CFO, and programmer).
    Based on the generated repository, we allow users to edit its content to achieve their final goal.
    \item \textbf{\codes Extended}: \codes includes two versions: prompt engineering (PE) and supervised fine-tuning (SFT). So, we develop VSCode extensions for these two versions separately based on CodeLlama $34$B. The extensions can generate an entire code repository through a three-layer sketch. 
    We allow the user to edit the generated content of each layer before proceeding to the next layer. Users can realize their requirements through interactions with VSCode extensions or manual modifications.
\end{itemize}

\subsection{Two Real-World Engineering Projects}
\label{sec:two_real_world_engineering_projects}

To evaluate the practicality of the above code assistants, we design two projects for participants to implement.
We provide brief descriptions of the projects below. For additional details, please visit the provided links.
\begin{itemize}
    \item \textbf{Gomoku (easy)}\footnote{https://github.com/NL2Code/CodeS/tree/main/projects/Gomoku}: This project is a two-player board game played on a $15$x$15$ grid. The objective of the game is to be the first player to get five of their pieces in a row, either horizontally, vertically, or diagonally. This game needs to be implemented using the \texttt{tkinter} library\footnote{https://docs.python.org/3/library/tkinter.html}.
    \item \textbf{Blog (hard)\footnote{https://github.com/NL2Code/CodeS/tree/main/projects/Blog}}: This project is a blog system. The features of this blog mainly include registration, login, blog management, and comment management. This blog requires to be implemented using the \texttt{Flask} framework\footnote{https://flask.palletsprojects.com}.
\end{itemize}

\begin{table*}[!t]
\caption{Results for five code assistants in two projects: \textbf{R./W.} for \underline{r}ead/\underline{w}rite time, \textbf{Req.} for the number of \underline{req}uesting assistants, \textbf{Code Prop.} for the auto-generated \underline{code} token \underline{prop}ortion, and \textbf{n/a} for projects not completed within $2.5$ days.}
\begin{center}
\scalebox{1.0}{
\rotatebox{0}{
\setlength{\tabcolsep}{3.5pt}
\begin{tabular}{l|ll|llll|llll} 
\toprule
\multirow{2}{*}{\textbf{Code Assistants}} & \multirow{2}{*}{\textbf{Type }} & \multirow{2}{*}{\begin{tabular}[c]{@{}l@{}}\textbf{Base}\\\textbf{Model}\end{tabular}} & \multicolumn{4}{c|}{\textbf{Gomoku (Easy)}}              & \multicolumn{4}{c}{\textbf{Blog (Hard)}}                  \\
                                &                                 &                                                                                        & \textbf{Time} & \textbf{R./W.} & \textbf{Req.} & \textbf{Code Prop.} & \textbf{Time} & \textbf{R./W.} & \textbf{Req.} & \textbf{Code Prop.}  \\ 
\hline\hline
Python Extended                 & Rule                            & -                                                                                      & 4.2h          & 1.4h/2.8h      & -             & -                   & 2.3d          & 0.7d/1.6d      & -             & -                    \\
ChatDev~Extended                & LLM                             & GPT-3.5                                                                                & 2.1h          & 1.5h/0.6h      & 103           & 67\%                & n/a           & n/a            & n/a           & n/a                  \\
GitHub Copilot                  & LLM                             & -                                                                                      & 1.6h          & 0.9h/0.7h      & 66            & 52\%                & 1.5d          & 0.8d/0.7d      & 316           & 45\%                 \\
\codes Extended (PE)             & LLM                             & CodeLlama 34B                                                                          & 2.2h          & 1.7h/0.5h      & 73            & 63\%                & 1.9d          & 1.3d/0.6d      & 263           & 50\%                 \\
\codes~Extended (SFT)            & LLM                             & CodeLlama 34B                                                                          & 1.9h          & 1.5h/0.4h      & 60            & 67\%                & 1.8d          & 1.3d/0.5d      & 247           & 52\%                 \\
\bottomrule
\end{tabular}
}
}
\label{tab:vscode_results}
\end{center}
\end{table*}

\subsection{Procedure of Our Empirical Study}
\label{sec:procedure_of_our_empirical_study}
For each of the five code assistants, we invite six participants to implement the two projects, three per project.
In total, we invite thirty participants.
For fairness, we ensure that these participants have similar backgrounds: both master students from the same university, both software engineering-related majors, and both with over two years of Python experience.
Before starting, we dedicate a $100$-minute training session, including explaining project requirements, demonstrating reference projects, and familiarizing with the usage of code assistants.
We also invite three Ph.D. students as proctors to ensure the process progressed orderly.
We systematically track and monitor their entire implementation process.
(1) We record the average completion time across three participants.
We also separately record the time they spent reading and writing code\footnote{More than ten seconds without any changes to the code is seen as reading, and vice versa for writing.}.
(2) We count the total number of requesting code assistants, and calculate the percentage of code tokens automatically generated by assistants.
(3) We count their common error types and invite all participants to evaluate the code assistants used.
(4) After all participants have completed their projects, we analyze the output code repositories in detail.
Specifically, we count the number of directories, files, lines, classes, and functions for the generated code repositories.
(5) Also, we use Pylint\footnote{https://pypi.org/project/pylint} to measure the code quality of the generated repositories (scoring from $0$ to $10$) and invite three proctors to manually review them (scoring from $0$ to $100$).

\subsection{Results}

\begin{figure}[!t]
    \small
    \centering
    \includegraphics[width=0.8\linewidth]{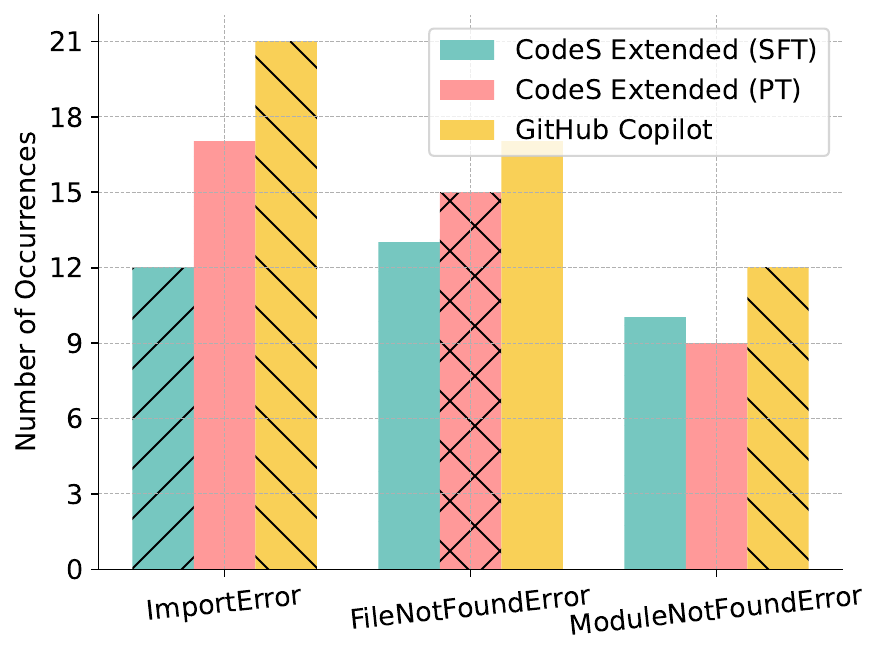}
    \caption{The number of occurrences of various error types under different code assistants.}
    \label{fig:common_errors}
\end{figure}

Table~\ref{tab:vscode_results} shows the performance of five code assistants on two real-world projects.
It can be observed that these LLM-based assistants significantly outperform the rule-based one.
For instance, Python Extended takes 4.2 hours to complete Gomoku, whereas GitHub Copilot and \codes Extended (SFT) complete the task in only 1.6 and 1.9 hours, respectively.
This observation underscores the practicability of LLM-based assistants in accomplishing real-world \nltorepo tasks.
Additionally, while ChatDev Extended achieves decent performance on Gomoku, it fails on the more intricate project Blog.
In contrast, \codes Extended demonstrates a more stable performance on both Gomoku and Blog projects, highlighting its effectiveness and reliability in addressing \nltorepo tasks with varied difficulties.
Compared with Copilot, users often spend far more time reading code than writing in the process of using \codes Extended.
Specifically, Copilot takes 0.9h/0.7h to read/write code, versus 1.5h/0.4h for \codes Extended (SFT).
After re-examining the implementation process, we discover that \codes Extended often generates more detailed code at once, which increases the workload of reading.
However, once the repository generated by \codes Extended is understood, users will be able to accomplish their requirements in a shorter time.
Regarding Copilot, its meticulous product design significantly enhances user experience~\cite{copilot,copilot2}, contributing to the reduced time required to complete the two projects, which is also empirically evidenced in the supplementary material (Section B).
This suggests that our plugin still has room for improvement in product design and other aspects.
To further compare the capability of \codes and Copilot to handle the dependencies within the repository, we count the occurrences of three dependency-related errors for different code assistants in Fig.~\ref{fig:common_errors}, including ImportError, FileNotFoundError, and ModuleNotFoundError. 
Compared to Copilot, \codes Extended raises fewer ImportError, FileNotFoundError, and ModuleNotFoundError, which highlights the advantages of \codes Extended in handling repository-level cross-file, and cross-module function invocations.

To evaluate the degree to which the code assistants adhere to human coding style, in Table~\ref{tab:vscode_project_details}, we compare the statistics between the code assistants' generated repositories and the reference repository.
It can be observed that \codes Extended (SFT) consistently receives the highest score on both the Gomoku and Blog projects. 
Compared with other code assistants, the repository structures generated by \codes Extended (PE) and (SFT) are more aligned with the reference repository, particularly for the hard project Blog.
These observations further highlight the practicality of \codes in assisting real-world software development tasks.

\begin{table}[!t]
\caption{Statistics of the repositories generated by five code assistants. \textbf{\#D./F./L.} means the number of \underline{d}irectories, \underline{f}iles, and \underline{l}ines; \textbf{\#c./f.} denotes the number of \underline{c}lass and \underline{f}unction; \textbf{Q.} represents the code \underline{q}uality score from Pylint; \textbf{S.} indicates grade \underline{s}cores from three proctors.}
\begin{center}
\scalebox{0.95}{
\rotatebox{0}{
\setlength{\tabcolsep}{2.0pt}
\begin{tabular}{l|llll|llll} 
\toprule
\multirow{2}{*}{\textbf{Tool }} & \multicolumn{4}{c|}{\textbf{Gomoku (Easy)}}                        & \multicolumn{4}{c}{\textbf{Blog (Hard)}}                                                                \\
                                & \textbf{\#D./F./L.} & \textbf{\#c./f.} & \textbf{Q.} & \textbf{S.} & \textbf{\textbf{\#D./F./L.}} & \textbf{\textbf{\#c./f.}} & \textbf{\textbf{Q.}} & \textbf{\textbf{S.}}  \\ 
\hline\hline
Reference Repository               & 0/2/101             & 3/12             & 8.4         & -           & 21/29/1,859                  & 28/119                    & 8.3                  & -                     \\ 
\hdashline
Python Extended                 & 0/1/94              & 2/9              & 5.7         & 74          & 13/20/1,562                  & 21/112                    & 6.1                  & 72                    \\
ChatDev~Extended                & 0/3/125             & 3/14             & 6.3         & 83          & n/a                          & n/a                       & n/a                  & n/a                   \\
GitHub Copilot                  & 0/1/104             & 2/12             & 6.8         & 77          & 13/21/1,615                  & 24/122                    & 6.5                  & 75                    \\
\codes~Extended (PE)             & 0/3/146             & 3/15             & 6.5         & 77          & 18/27/1,834                  & 25/117                    & 6.4                  & 73                    \\
\codes~Extended (SFT)            & 0/2/133             & 4/17             & 7.2         & 84          & 19/26/1,774                  & 27/120                    & 6.9                  & 77                    \\
\bottomrule
\end{tabular}
}
}
\label{tab:vscode_project_details}
\end{center}
\end{table}

\section{Related Work}

The task of \nltorepo necessitates long inputs as well as long structured outputs, presenting new challenges:

\textbf{\nltorepo requires handling long contexts.} 
Most of the existing language models support context window sizes ranging from 4k to 32k~\cite{codex,codegen,starcoder,codegeex,incoder,codet5,cert,pangucoder,alphacode}.
Some support longer context up to 128K (Gemini), 200K (Claude 2.1)~\cite{deepseekcoder,codellama,starcoder2,gpt4}.
Rotary Position Embedding (ROPE)~\cite{rope} or the technique that concatenates multiple windows~\cite{paraCxtEncoding, 400k} can extend the context window size effectively.  
However, even with the largest context window size, it is not sufficient for repository-level tasks due to the large size of code bases, documentation, and discussions.

\textbf{\nltorepo requires efficient instruction fine-tuning.}
Instruction fine-tuning has been proved to be effective for multiple downstream code tasks~\cite{wizardcoder, pangucoder2}. 
It is also used to enhance a wide variety of base models, such as WizardCoder~\cite{wizardcoder},
CodeLlama-Instruct~\cite{codellama}, and DeepSeek-Coder-Instruct~\cite{deepseekcoder}.
For complex tasks with long inputs and outputs. It not only has a longer fine-tuning time for one data instance but also needs more tuning data. 
It is usually not affordable to do a full parameter tuning.
To reduce cost, LongLora~\cite{longLora} proposes sparse local attention which can finish fine-tuning for 32K window in around 11 hours with 8 A100 GPUs.

\textbf{Benchmarks for complex coding tasks.} 
Benchmarks for code generation are designed with different levels of context information but also target different granularity of code snippets.
CoNaLa~\cite{conala} targets line completion. 
HumanEval~\cite{codex}, APPs~\cite{apps}, CONCODE~\cite{concode}, and CoderEval~\cite{coderEval} focus function generation.
Moreover, RepoEval~\cite{repoeval}, CrossCodeEval~\cite{crosscodereval}, and RepoBench~\cite{repobench} focus on cross-file line/function-level code generation.
Beyond lines and functions, class-level~\cite{classeval}, and patch-level~\cite{swebench} generation is drawing more and more attention.
This paper is the first to propose a repository-level benchmark \sketcheval for evaluating the \nltorepo tasks.

\textbf{Towards complex coding tasks with LLM-centered systems.}
Tasks like \nltorepo may be too hard for a single LLM to solve.  
Iterative generation with feedback from tools or humans seems to be mandatory.
It may also require the capability of reasoning and the decomposition of tasks. 
Like Parsel~\cite{parsel}, it reasons about code in the form of call graphs.
Besides reasoning, feedback from test will be critical for correctness. Executing tests can identify the wrong generation in the early stage~\cite{alphaCodim,parsel}.
Still, iterating only with tools or a "single-minded" LLM is not sufficient. 
New perspectives are also important for problem solving, either provided by different LLMs or humans, like ChatDev~\cite{chatdev} utilizes communications between different agents, ANPL~\cite{anpl} interacts with humans.
The most recent work Devin~\cite{devin} shows the potential of fully automated software development. 
This paper uses the sketch idea to empower LLMs to address the complex \nltorepo task, achieving decent performance.

\section{Threats to Validity}

\textbf{Internal Threats.}
The internal threats mainly lie in four aspects:
(1) The \codes comprises three modules (i.e., \reposketcher, \filesketcher, and \sketchfiller), which respectively correspond to the three layers of the code repository architecture.
However, the structural hierarchy within the repository is more intricate than these three layers, which has not been fully taken into consideration.
We would like to explore how to implement \nltorepo via a more fine-grained sketch.
(2) The invocation relationships in real-world engineering repositories are typically complex. 
Thus, ensuring the consistency and accuracy of invocation relationships among multiple modules in \codes poses a considerable challenge.
In our supplementary material (Section C), we experiment with sequentially generating files based on cross-file definitions and references, yielding a performance gain, particularly for those complex repositories.
We believe that introducing techniques in program static analysis such as call graph analysis~\cite{cga} can further alleviate this issue.
(3) Although we have decomposed the \nltorepo task with the sketch and enabled ROPE~\cite{rope} during fine-tuning, the limited long-context capability of LLMs remains a constraint for overly large repositories.
A possible solution is to highly condense information using LLMs' summarization capabilities, which remains to be explored in the future.
(4) Our approach, which is based on the multi-layer sketch, may be susceptible to cascading errors.
Fortunately, human intervention can mitigate this issue by supervising the generated results of each layer.
The experimental results presented in Section~\ref{sec:rq3_the_practicality_of_codes} can also prove the practical potential of \codes.

\textbf{External Threats.}
Our external threats encompass three aspects:
(1) Despite our best efforts to gather the most current open-source repositories (2023.08.01$\sim$) for evaluation, some recently released models, such as StarCoder2 ($\sim$2023.09.06), still possibly have seen them during pre-training, due to the rapid evolution of LLMs.
(2) Due to the task's multifaceted nature, it is challenging to devise a metric that is completely fair and universally accepted.
However, other than benchmark-based experiments evaluated with \sketcheval, we also conduct feedback-based experiments to provide a more diverse and comprehensive evaluation.
(3) This paper focuses on Python-version general code repositories.
In future work, we will adopt the idea of \codes to more languages (e.g., Java and C++) and more vertical programming tasks (e.g., web development and robot control).
Interestingly, \codes's Python version can be generalized to other languages, with details in the supplementary materials (Section D).

\section{Conclusion}
This paper proposes a new software engineering task, namely natural language to code repository (\nltorepo).
To address this task, we propose a framework called \codes.
It can decompose a complex \nltorepo task into multiple sub-tasks via a multi-layer sketch, and then solve them layer by layer.
To evaluate \codes on the \nltorepo task, we craft a new benchmark \sketcheval, 
and provide an evaluation metric \sketchbleu.
Furthermore, we develop a VSCode plugin for \codes and invite $19$ participants to conduct a comprehensive empirical study.
Extensive experiments have proved the effectiveness and practicality of \codes.

\section*{Acknowledgments}
This research was supported by the National Key Research and Development Program of China, under Grant No. 2022ZD0120201 - ``Unified Representation and Knowledge Graph Construction for Science Popularization Resources''.

\clearpage

\textbf{~~~~~~~~~~~~~~~Supplementary Material}

\subsection{A detailed specification of \sketchbleu}
\label{apx:sketchbleu_specification}

Resembling CodeBLEU~\cite{codebleu}, \sketchbleu is a weighted combination of four sub-parts ($BLEU'$, $BLEU_{weight}'$, $Match_{struc}$ and $Match_{df}'$), in place of the original four sub-parts of CodeBLEU ($BLEU$, $BLEU_{weight}$, $Match_{ast}$ and $Match_{df}$).
To transfer the function-level metric CodeBLEU to a new repository-level metric SketchBLEU, we make several modification to the original sub-parts of the CodeBLEU metric.
Specifically, we make four main modifications:
\textbf{First}, we replace the single code snippet used to calculate the original $BLEU$ and $BLEU_{weight}$ with a concatenation of all the code files in the repository.
\textbf{Second}, we replace the BLEU brevity penalty with a less sensitive one because the size of repositories designed to meet the same requirements can differ markedly.
The modified brevity penalty can be formalized as:
\begin{equation}
    BP_{\frac{c}{r}}' = 
    \begin{cases}
        1 \qquad\qquad\quad\ if\ 2\cdot c > r \\
        \frac{1}{1+\ln{r}-\ln{2\cdot c}} \quad if\ 2\cdot c \leq r
    \end{cases}
\end{equation}
\textbf{Third}, we replace the ASTs used to calculate the original $Match_{ast}$ with a structural tree combining both directory structure and source code syntax, thus deriving $Match_{struc}$.
Also, instead of counting the fully matched sub-trees as the original $Match_{ast}$ does, we extract n-hop sub-trees, truncating the sub-trees to limit its depth.
The reason behind this modification is that, compared to function-level codes, repository have a multi-layered structure, hard to be evaluated based on a full-depth sub-tree.
\textbf{Fourth}, we extend the original function-to-function dataflow match metric $Match_{df}$ to a repository-to-repository one, namely $Match_{df}'$:
\begin{equation}
\begin{split}
    Match_{df}' =
    \begin{cases}
       BP_{\frac{|Hyp|}{|Ref|}}'\cdot\frac{\texttt{MWBM}}{|Hyp|}\quad if |Ref| > |Hyp|\\
       BP_{\frac{|Ref|}{|Hyp|}}'\cdot\frac{\texttt{MWBM}}{|Ref|}\quad if |Ref| \leq |Hyp|
    \end{cases}
\end{split}
\end{equation}
where $\texttt{MWBM}$ denotes the maximum bipartite weighted matching of a complete bipartite graph with $(Ref,Hyp)$ as the partition and $Match_{df}$ as the weight function;
$Ref$ and $Hyp$ respectively correspond to functions in the reference repository and the prediction repository.
We also use the modified brevity penalty $BP'$ to measure the size gap between the function sets of the reference repository and the prediction repository.
For more details, please refer to the implementation of \sketchbleu, which can be found in our open-source repository\footnote{https://github.com/NL2Code/CodeS/tree/main/validation/evaluation\_scripts}.

\subsection{Effectiveness Evaluation from Participants}
\label{sec:apx_effectiveness_evaluation_from_participants}

\begin{figure}[!t]
    \small
    \centering
    \includegraphics[width=0.65\linewidth]{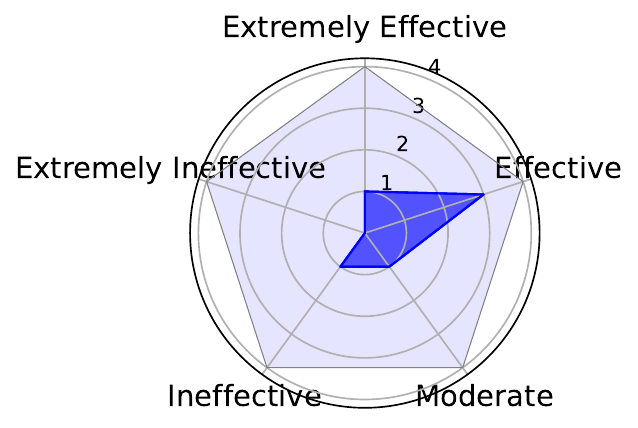}
    \caption{User survey of $6$ participants using \codes Extended.}
    \label{fig:survey_effectiveness_evaluation}
\end{figure}

After completing the project, we conduct a user survey with six participants who used the \codes extended.
The survey results in Fig.~\ref{fig:survey_effectiveness_evaluation} show that most participants deem the plugin we developed to be effective, further proving the effectiveness and practicality of \codes.
Meanwhile, we also note a participant who deem \codes Extended ineffective.
Further inquiry with this participant reveal that \codes Extended falls short in product design compared to the commercial product GitHub Copilot.
This conclusion is also reflected in Section~\ref{sec:rq3_the_practicality_of_codes}, suggesting significant room for improvement in product design for our developed VSCode plugin.

\subsection{The Impact of The Generation Order of Files on \codes}
\label{sec:apx_order_generation}
\codes does not distinguish the generation order of files.
We would like to generate files in accordance with the sequence determined by their invocation relationships.
Taking \ding{193} of Fig.~\ref{fig:codes_three_modules} (a) as an example, 
we can observe that ``mongio.py'' invokes ``settings.py''.
So, we generate ``settings.py'' before generating ``mongio.py''; the same principle applies to the remaining files.
Results in Table~\ref{tab:apx_ordered_generation} indicate that ordered generation leads to a performance improvement.
Compared to easy repositories, this strategy exhibits a more pronounced advantage on hard ones.
This is primarily due to that more hard repositories often involve intricate invocation relationships between files, which are less prevalent in easier ones.

\begin{table}[!t]
\caption{\sketchbleu of \codes (SFT) using inter-file topological ordering.}
\begin{center}
\scalebox{1.0}{
\rotatebox{0}{
\setlength{\tabcolsep}{3.5pt}
\begin{tabular}{l|llll} 
\toprule
\multirow{2}{*}{\textbf{Method}} & \multicolumn{4}{c}{\textbf{SketchEval}}                                          \\
                                 & \textbf{Easy (5)} & \textbf{Middle (8)} & \textbf{Hard (6)} & \textbf{All (19)}  \\ 
\hline\hline
\codes (CodeLlama 7B)             & 55.96             & 56.11               & 50.78             & 54.39              \\
~ +Ordered Generation            & 56.67             & 56.32               & 51.00             & 54.73              \\ 
\cdashline{1-1}
\codes (CodeLlama 13B)            & 58.24             & 57.25               & 56.35             & 57.23              \\
~ +Ordered Generation            & 58.15             & 57.66               & 58.46             & 58.04              \\ 
\cdashline{1-1}
\codes (CodeLlama 34B)            & 65.14             & 64.24               & 60.56             & 63.31              \\
~ +Ordered Generation            & 65.36             & 64.63               & 61.63             & 63.87              \\
\bottomrule
\end{tabular}
}
}
\label{tab:apx_ordered_generation}
\end{center}
\end{table}

\subsection{Generalization of \codes}
\label{sec:generalization_of_codes}
We are intrigued to explore whether \codes trained on Python repositories can be generalized to other languages.
Therefore, we generate repository sketch for Java, C++, and Go projects using \codes's \reposketcher\footnote{We craft an evaluation set of $19$ repositories for each language, following the same process outlined in Section~\ref{sec:repository_collection_of_benchmark}. These repositories can be seen at https://github.com/NL2Code/CodeS/tree/main/validation/multilingual-repos.}.
In Table~\ref{tab:java_performance}, we evaluate two versions of \reposketcher: prompt engineering and supervised fine-tuning.
We observe that after fine-tuning on Python, \reposketcher also boosts performance on other languages.
Such the observation illustrates \codes can achieve generalization across multiple programming languages.

\begin{table}[!t]
\caption{Repository Sketch's BLEU of \reposketcher (fine-tuned on Python) on different languages.}
\begin{center}
\scalebox{1.0}{
\rotatebox{0}{
\setlength{\tabcolsep}{2.5pt}
\begin{tabular}{l|ll} 
\toprule
\multirow{2}{*}{\begin{tabular}[c]{@{}l@{}}\textbf{Evaluation}\\\textbf{Languages}\end{tabular}} & \multicolumn{2}{c}{\textbf{\reposketcher of \codes (CodeLlama $13$B)}}                       \\
                                                                                                 & \textbf{Prompt Engineering} & \textbf{Supervised Fine-Tuning on Python}  \\ 
\hline\hline
Java                                                                                             & 46.25                       & 57.36                                      \\
C++                                                                                              & 36.68                       & 50.14                                      \\
Go                                                                                               & 62.88                       & 71.57                                      \\
\bottomrule
\end{tabular}
}
}
\label{tab:java_performance}
\end{center}
\end{table}

\subsection{Details of Our Empirical Study}
\label{apx:details_of_our_empirical_study}

(1) For the Blog project, we provide participants with detailed database SQL files and migrations\footnote{https://flask-migrate.readthedocs.io}, and clarify all database fields for them.
(2) In the design of \codes Extended, we generate non-Python content by harnessing \codes's intrinsic capabilities via prompt engineering.
(3) The three proctors score a code repository according the completion of its features and the quality of its code.
(4) In Table~\ref{tab:three_phase_results}, the time calculation only included the effective time using the code assistant.
(5) We invite a total of thirty participants, none of whom are authors.

\end{document}